\documentclass[letterpaper, 10 pt, conference]{ieeeconf}
\IEEEoverridecommandlockouts 
\usepackage{graphics}
\usepackage{amsmath}
\usepackage{amssymb}
\usepackage{xcolor}
\usepackage{graphicx}
\usepackage{wasysym}
\usepackage{ulem} 
\usepackage{tcolorbox}
\usepackage{booktabs}
\usepackage{siunitx}
\usepackage{textcomp}
\usepackage{pifont}

\newcolumntype{N}{>{\scriptsize}l}

\definecolor{matlab1}{rgb}{0.0000, 0.4470, 0.7410}
\definecolor{matlab2}{rgb}{0.8500, 0.3250, 0.0980}
\definecolor{matlab3}{rgb}{0.9290, 0.6940, 0.1250}

\definecolor{matlab5}{rgb}{0.0000, 0.7059, 0.0000}
\definecolor{matlab7}{rgb}{0.7059, 0.0000, 0.0000}

\newcommand{\checkNo}{\hfill \textcolor{matlab2}{\ding{55}} \hfill}
\newcommand{\checkYes}{\hfill \textcolor{matlab5}{\ding{51}} \hfill}

\def\diag{\operatorname{diag}}

\def\transpose{^\top}

\usepackage{tikz}
\newcommand\copyrighttext{%
    \footnotesize 
    \color{gray}
    \textcopyright  
    2024
    \ IEEE. Personal use of this material is permitted.  Permission from IEEE must be obtained for all other uses, in any current or future media, including reprinting/republishing this material for advertising or promotional purposes, creating new collective works, for resale or redistribution to servers or lists, or reuse of any copyrighted component of this work in other works.
}

\title{\LARGE \bf
On the Disentanglement of Tube Inequalities in\\
Concentric Tube Continuum Robots
}

\author{Reinhard M. Grassmann$^{*, 1}$, Anastasiia Senyk$^{*, 1, 2}$, and Jessica Burgner-Kahrs$^{1}$, \textit{Senior Member, IEEE}
\thanks{
We acknowledge the support of the Natural Sciences and Engineering Research Council of Canada (NSERC), [RGPIN-2019-04846] as well as the Canada Foundation for Innovation and Ontario Research Fund [Project \#40110].
}
\thanks{
$^{1}$Continuum Robotics Laboratory, Department of Mathematical and Computational Sciences, University of Toronto, Mississauga, ON L5L 1C6, Canada. {\tt\small reinhard.grassmann@utoronto.ca}
}
\thanks{
$^{2}$Faculty of Applied Science, Ukrainian Catholic University, Lviv, Ukraine.
}
\thanks{
$^{*}$Equal contribution.
}
}

\begin{document}

\maketitle
\thispagestyle{empty}
\pagestyle{empty}

\begin{abstract}

Concentric tube continuum robots utilize nested tubes, which are subject to a set of inequalities.
Current approaches to account for inequalities rely on branching methods such as if-else statements.
It can introduce discontinuities, may result in a complicated decision tree, has a high wall-clock time, and cannot be vectorized.
This affects the behavior and result of downstream methods in control, learning, workspace estimation, and path planning, among others.

In this paper, we investigate a mapping to mitigate branching methods.
We derive a lower triangular transformation matrix to disentangle the inequalities and provide proof for the unique existence.
It transforms the interdependent inequalities into independent box constraints.
Further investigations are made for sampling, control, and workspace estimation.
Approaches utilizing the proposed mapping are at least 14 times faster (up to 176 times faster), generate always valid joint configurations, are more interpretable, and are easier to extend.

\end{abstract}

\begin{tikzpicture}[remember picture,overlay]
        \node[anchor=south,yshift=10pt] at (current page.south) {\parbox{\dimexpr0.75\textwidth-\fboxsep-\fboxrule\relax}{\copyrighttext}};
\end{tikzpicture}


\vspace{-1.25em} 
\section{INTRODUCTION}

A concentric tube continuum robot (CTCR) is composed of several nested concentric tubes being pre-curved, and super-elastic \cite{SearsDupont_IROS_2006, WebsterOkamuraCowan_IROS_2006}.
The inherent compliant and flexible nested tubes are translated and rotated with respect to each other to change the manipulator's shape.
Different tube materials can achieve this elastic interaction between tubes.
Such robots typically comprise two or three metal alloy tubes, usually nitinol \cite{NwaforRabenorosoa_et_al_TRO_2023}. 
CTCR robot's distinct feature lies in the nonlinear dynamics resulting from the elastic interplay between these tubes.
On top of this, from a mechanical and modeling point of view, the use of nested tubes imposes tube length constraints formulated as a set of two inequalities.
Unfortunately, little attention is paid to mitigating the effects of these inequalities.

\begin{figure}[t!]
	\vspace*{2mm}
	\centering
    \includegraphics[width=\columnwidth]{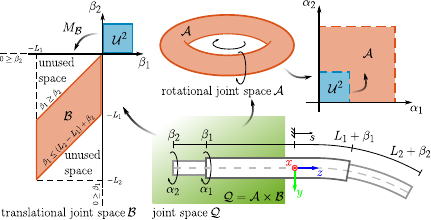}
	\vspace*{-1.5em}
	\caption
    {
        Characteristics of the joint space of a \SI{4}{dof} CTCR with two nested tubes. 
        While tube rotations $\alpha_i$ form a torus, therefore, can be represented as square, the tube translations $\beta_i$ result in a parallelogram.
        This parallelogram is bounded by the inequalities.
        Considering this geometric insight, it follows that an affine transformation $M_\mathcal{B}$ can be used to map a square $\mathcal{U}^2$ to the parallelogram $\mathcal{B}$.
        The geometric insight extends to $N$ tubes, and it should be clear that the transformed space, i.e., $\mathcal{U}^N$ for $N$ tubes, is more desirable.
    }
	\label{fig:catchy_image}
	\vspace*{-1.5em}
\end{figure}

Consider a CTCR with two tubes being subject to the two inequalities $0 \geq\ \beta_1 \geq \beta_2$ and $0 \leq L_1 + \beta_1 \leq L_{2} + \beta_{2}$.
As illustrated in Fig.~\ref{fig:catchy_image}, the translational joint space $\mathcal{B}$ forms a parallelogram.
Both inequalities span the edges of the translational joint space $\mathcal{B}$.
From this observation, we can use rotation, scaling, and shearing to obtain
\begin{align}
    \underbrace{
	\begin{bmatrix}
		1 & 0\\
		1 & 1
	\end{bmatrix}
    }_\text{shearing}
    \underbrace{
	\begin{bmatrix}
		L_1 & 0\\
		0 & L_2  - L_1
	\end{bmatrix}
    }_\text{scaling}
    \underbrace{
	\begin{bmatrix}
		-1 & 0\\
		0 & -1
	\end{bmatrix}
    }_\text{rotation}
    \underbrace{
    \begin{bmatrix}
        \beta_{1, \mathcal{U}}\\
        \beta_{2, \mathcal{U}}
    \end{bmatrix}
    }_{\substack{\in \, \mathcal{U}^2}}
    = 
    \underbrace{
        \begin{bmatrix}
            \beta_{1}\\
            \beta_{2}
        \end{bmatrix}
    }_{\substack{\in \, \mathcal{B}}}.
    \nonumber
\end{align}
We introduced this type of transformation in \cite{GrassmannBurgner-Kahrs_RSS_2019}.
For learning the kinematics of a CTCR task, we show in \cite{GrassmannBurgner-Kahrs_RSS_2019} that the accuracy and convergence are significantly improved by using this simple yet effective transformation to decorrelate the joint space.
As of now, this approach has been used in publications \cite{GrassmannBurgner-Kahrs_RSS_2019, GrassmannBurgner-Kahrs_et_al_IROS_2022, IyengarStoyanov_et_al_RA-L_2023} on machine learning. Yet, we are confident that disentanglement will also be useful for other research fields in the continuum robotics research community.
Furthermore, looking at applications beyond CTCRs, the disentanglement of inequalities is also useful for continuum robot types with variable lengths and cranes with prismatic joints.
For instance, a tendon-driven continuum robot's variable segments can be achieved using tubes \cite{NguyenBurgner-Kahrs_IROS_2015}.

More importantly, one or both inequalities are usually neglected for approaches purely evaluated in simulation, e.g., \cite{BoushakiPoignet_ICRA_2014, FellmannKashiBurgner-Kahrs_MI_2015, SabetiaDrake_et_al_RA-L_2019}.
Observing the proliferation of prototypes \cite{GrassmannBurgner-Kahrs_et_al_RSS_WS_2020} poses a risk for translating the results on physical hardware due to the need for considering both inequalities.
Furthermore, if both inequalities are considered, methods in the literature rely on branching methods.
Branching methods such as if-statements can be implemented straightforwardly.
However, they can introduce discontinuities, have high wall-clock time, and are poorly vectorizable.
Consequently, methods and approaches using branching methods may not be suitable for controlling, learning, optimization, workspace estimation, and path planning.
Especially computationally expensive applications or applications considering small time steps would suffer from a high wall-clock time.

In this paper, we expand on our previously proposed affine transformation matrix \cite{GrassmannBurgner-Kahrs_RSS_2019, GrassmannBurgner-Kahrs_et_al_IROS_2022} by providing a proof for its uniqueness, discussing its theoretical foundation, and showing its effectiveness in applications beyond learning, i.e., sampling, workspace estimation for concentric tube continuum robots, and control synthesis.

\section{Disentanglement of Beta}

In this section, we recap the affine transformation matrix to disentangle the translational joint space as illustrated in Fig.~\ref{fig:catchy_image}.
While this transformation is proposed in our previous work \cite{GrassmannBurgner-Kahrs_RSS_2019, GrassmannBurgner-Kahrs_et_al_IROS_2022}, we present more theoretical insights and identify patterns by explicitly generalizing to $N$ tubes.
On top of this, we prove the uniqueness of the affine transformation and show that only one such mapping exists.

\subsection{Affine Transformation Matrix}

Due to the use of nested tubes, both inequalities given by
\begin{align}
	0 \geq&\ \beta_1 \geq \beta_2 \geq \cdots \geq \beta_N \quad\text{and}\label{eq:inequality_beta}\\
	0 \leq& L_1 + \beta_1 \leq L_{2} + \beta_{2} \leq \cdots \leq L_N + \beta_N, \label{eq:inequality_L}
\end{align}
have to be satisfied by each translation $\beta_i$.
Here, the constant $L_i$ is the length of the $i\textsuperscript{th}$ tube.
The interdependencies caused by \eqref{eq:inequality_beta} and \eqref{eq:inequality_L} result from the fact that none of the distal ends of inner tubes should be inserted inside the respective outer tube.
This can be derived both from a theoretical point of view \cite{RuckerJonesWebster_TRO_2010} and from a practical point of view, e.g., to avoid play between adjacent tubes, which can cause hysteresis \cite{Greiner-PetterSattel_SMS_2017}.
Furthermore, \eqref{eq:inequality_beta} can prevent collision between carriages, while \eqref{eq:inequality_L} can prevent collision between the outer tube and gripper or attached sensors \cite{GrassmannBurgner-Kahrs_et_al_IROS_2022}.

To avoid the evaluation of the inequalities \eqref{eq:inequality_beta} and \eqref{eq:inequality_L}, the translations $\beta_i$ can be transformed such that the new joint space representation $\beta_{i, \mathcal{U}}$ are decorrelated.
For the sake of compactness, we summarize all $N$ translations $\beta_i$ and $\beta_{i, \mathcal{U}}$ in a column matrix $\boldsymbol{\beta}$ and $\boldsymbol{\beta}_{\mathcal{U}}$, respectively. 
For the transformation $\boldsymbol{\beta} = \boldsymbol{M}_\mathcal{B}\boldsymbol{\beta}_{\mathcal{U}}$, the mapping matrix given by
\begin{align}
    %
    \small 
	\boldsymbol{M}_\mathcal{B}
	=
	\begin{bmatrix}
		-L_1	&   0           &	0          & \cdots & 0\\
		-L_1    &   L_1 - L_2   &	0          & \cdots & 0\\
		-L_1    &	L_1 - L_2   &	L_2 - L_3  & \cdots & 0\\
		\vdots  &	\vdots   &	\vdots  & \ddots & \vdots\\
		-L_1    &	L_1 - L_2   &	L_2 - L_3  & \cdots & L_{N-1} - L_N\\
	\end{bmatrix}
	\label{eq:M_B}
\end{align}
is defined for $N$ tubes and follows the used convention in \cite{GrassmannBurgner-Kahrs_et_al_IROS_2022} and shown in Fig.~\ref{fig:catchy_image}, i.e., most inner tube $i = N$ and outermost tube $i = 1$.
The determinant of \eqref{eq:M_B} is  
\begin{align}
    \det \boldsymbol{M}_\mathcal{B} = \left(-1\right)^N L_1\prod_{i=2}^N \left(L_{i} - L_{i-1}\right).
	\label{eq:M_B_det}
\end{align}
The factors of the product are the eigenvalues, and the sign of the determinant changes depending on if $N$ is odd or even.
More importantly, since $0 < L_{i-1} < L_i$, the inverse of \eqref{eq:M_B} always exists.
The inverse of the lower triangular matrix is
\begin{flalign}
    %
    \small 
    \hspace*{-2.5mm}
	\boldsymbol{M}_\mathcal{B}^{-1}
    \!=\!
	\begin{bmatrix}
		\dfrac{1}{-L_1}	&   0           &	0          & \cdots & 0\\[1em]
		\dfrac{1}{L_2 \!-\! L_1}    &   \dfrac{1}{L_1 \!-\! L_2}   &	0          & \cdots & 0\\[1em]
		0    &	\dfrac{1}{L_3 \!-\! L_2}   &	\dfrac{1}{L_2 \!-\! L_3}  & \cdots & 0\\[1em]
		\vdots  &	\vdots   &	\vdots  & \ddots & \vdots\\[1em]
		0    &	0   &	0  & \cdots & \dfrac{1}{L_{N\!-\!1} \!-\! L_N}\\
	\end{bmatrix}
    \hspace*{-2.5mm}
	\label{eq:M_B_inverse}
\end{flalign}
being a matrix with nonzero values in the lower diagonal and main diagonal.
By computing $\boldsymbol{M}_\mathcal{B}^{-1}\boldsymbol{\beta}$, all entries of $\boldsymbol{\beta}_{\mathcal{U}}$ are in the interval $\left[0, 1\right]$.

To enforce $\beta_{i, \mathcal{U}} \in \left[-1, 1\right]$, the matrix \eqref{eq:M_B} is scaled and shifted.
Removing the homogeneous extension leads to
\begin{align}
    \boldsymbol{\beta}
	=
	\begin{bmatrix}
		\dfrac{1}{2}\boldsymbol{M}_\mathcal{B}    & \dfrac{1}{2}\boldsymbol{M}_\mathcal{B}\cdot\boldsymbol{1}_{N \times 1}
	\end{bmatrix}
	\boldsymbol{\beta}_{\mathcal{U}}.
	\label{eq:M_B4x4}
\end{align}
Inverting the homogeneous extension of \eqref{eq:M_B4x4} leads to
\begin{align}
	\boldsymbol{\beta}_{\mathcal{U}}
	=
	\begin{bmatrix}
		2\boldsymbol{M}_\mathcal{B}^{-1}    & -\boldsymbol{1}_{N \times 1}
	\end{bmatrix}
	\boldsymbol{\beta}
	\label{eq:M_B4x4_inverse}
\end{align}
without the homogeneous extension, where all $\beta_{i, \mathcal{U}}$ are automatically scaled between $-1$ and $1$ as well as unit-less and orthogonal.
The previously interdependent translational parameters $\beta_i$ subject to \eqref{eq:inequality_beta} and \eqref{eq:inequality_L} are now disentangled.
In fact, the inequalities become independent $N$ box constraints, i.e., $-1 \leq \beta_{i, \mathcal{U}} \leq 1$.

\subsection{Considering Minimal Length}

It is worth noting that $L_i$ is not necessarily the physical length of the tube. 
We follow an approach that is similar to the configuration approach in \cite{Lozano-Perez_TOC_1983} and capture additional mechanical constraints resulting from the actuation unit.
This gives the advantage that $\beta_i \in \left[\beta_{i, \text{min}}, 0\right]$, where $\beta_{i, \text{min}} = -L_i$.
In order to consider a safety margin for sensors \cite{GrassmannBurgner-Kahrs_et_al_IROS_2022}, the minimum displacement denoted by $\beta_{i, \text{min}}$ can be further restricted by considering a margin $L_{i, \text{margin}}$.
We can define 
\begin{align}
    \beta_{i, \text{min}} = -\left(L_i - \sum_{k=i}^N L_{k, \text{margin}}\right) = - L_i^*
    \label{eq:beta_min}
\end{align}
as the minimum displacement, where the restriction is applied recursively.
Afterwards, $L_i$ in \eqref{eq:M_B} and \eqref{eq:M_B_inverse} are substituted by $L_i^* = L_i - \sum_{k=i}^N L_{k, \text{margin}}$ to obtain $\boldsymbol{M}_{\mathcal{B}^*}$ and $\boldsymbol{M}_{\mathcal{B}^*}^{-1}$, where the star notation indicates the consideration of $L_i^*$.

\subsection{Uniqueness of the Low Triangular Transformation Matrix}
\label{sec:uniqueness_proof}

Now, we illustrate the derivation of \eqref{eq:M_B}.
The main idea is to first independently select a value for $\beta_{k}$ of the $k\textsuperscript{th}$ tube.
Consequently, values for the adjoint tubes, i.e., $(k-1)\textsuperscript{th}$ tube and $(k+1)\textsuperscript{th}$ tube, depend on the chosen $\beta_k$.
This step is repeated until the most inner and outermost tubes are reached and all values are chosen.
Consequently, for a given set of $N$ different tubes, it is possible to find $N$ different starting points for this approach and eventually $N$ different affine transformation matrices.
However, only one of them satisfies both inequalities \eqref{eq:inequality_beta} and \eqref{eq:inequality_L}, which is \eqref{eq:M_B}.

Starting with the outermost tube, we see the pattern
\begin{align}
    %
    %
    %
    \beta_1 &= -L_1 \beta_{1, \mathcal{U}}\label{eq:beta_i_induction_most_outer_first_step}\\
    \beta_2 &= \beta_1 - (L_2 - L_1) \beta_{2, \mathcal{U}} = \beta_1 + (L_1 - L_2) \beta_{2, \mathcal{U}}
    \nonumber
    \\
    \vdots
    \nonumber
    \\
    \beta_i &= \beta_{i-1} + (L_{i-1} - L_{i}) \beta_{i, \mathcal{U}} \label{eq:beta_i_induction_most_outer}
\end{align}
emerging, where $\beta_{i, \mathcal{U}} \in \mathcal{U}\left[0, 1\right]$ is a uniform distributed random variable.
Introducing $\beta_0 = 0$ and $L_0 = 0$, the first sampling step \eqref{eq:beta_i_induction_most_outer_first_step} is considered in pattern \eqref{eq:beta_i_induction_most_outer}, which can be used as base case for an induction proof.
Furthermore, looking at the lower bound, i.e., $\beta_{i, \mathcal{U}} \geq 0$, we can obtain a short hand for the inequality \eqref{eq:inequality_beta} being $\beta_i \leq \beta_{i-1}$.
In addition, considering the upper bound, i.e., $\beta_{i, \mathcal{U}} \leq 1$, the pattern  \eqref{eq:beta_i_induction_most_outer} simplifies to $\beta_i \leq \beta_{i-1} +  (L_{i-1} - L_{i})$.
After rearranging leading to $\beta_i + L_{i} \leq \beta_{i-1} +  L_{i-1}$, the other inequality \eqref{eq:inequality_L} can be distilled.
The result shows that \eqref{eq:beta_i_induction_most_outer} considers both inequalities, including the case for $i = 0$.
Using induction, the critical fact that the pattern \eqref{eq:beta_i_induction_most_outer} is always negative, i.e., $\beta_i \leq 0$, is reinforced as the base case \eqref{eq:beta_i_induction_most_outer_first_step} is negative and $L_{i-1} - L_{i} \leq 0$.

Forward substitution is used to obtain $\boldsymbol{M}_\mathcal{B}$.
From \eqref{eq:beta_i_induction_most_outer}, it is clear that this leads to a lower triangular matrix, where the entries of a column are either zero or the coefficient of $\beta_{i, \mathcal{U}}$ in \eqref{eq:beta_i_induction_most_outer}.
More importantly, the subsequently derived $\boldsymbol{M}_\mathcal{B}$ given by \eqref{eq:M_B} takes both inequalities \eqref{eq:inequality_beta} and \eqref{eq:inequality_L} into account.

Now, we look into an alternative pattern.
For this purpose, one can start with the most inner tube, i.e., $\beta_N = -L_N \beta_{N, \mathcal{U}}$, as the base case.
The emerging pattern is $\beta_i = \beta_{i+1} + (L_{i+1} - L_{i}) \beta_{i, \mathcal{U}}$.
This pattern leads to an upper triangular matrix.
Descending to $i = 0$, the pattern results in $\beta_{1} =  - L_1 \beta_{0, \mathcal{U}}$, where $\beta_{0, \mathcal{U}}$ is undefined.
More critical, it is not ensured that $\beta_i$ is always negative.
To show this, we attempt a proof by contradiction.
Assuming the emerging pattern is positive, i.e., $0 \leq \beta_{i+1} + (L_{i+1} - L_{i}) \beta_{i, \mathcal{U}}$, for the $i = N - 1$ case, we get $0 \leq \beta_{N} + (L_{N} - L_{N-1}) \beta_{N, \mathcal{U}}$.
Rearranging and substituting the base case, this yields $\beta_{N, \mathcal{U}} \leq (L_{N} - L_{N-1})/L_N \beta_{N-1, \mathcal{U}}$.
Hence, the attempt fails, and a condition between the random variables $\beta_{N, \mathcal{U}}$ and $\beta_{N-1, \mathcal{U}}$ exists, where $\beta_N$ can be positive.
Note that, for \eqref{eq:beta_i_induction_most_outer}, this would lead to a contradiction since the right side of \eqref{eq:beta_i_induction_most_outer} is always negative.
Therefore, starting with the inner tube does not lead to a valid mapping respecting \eqref{eq:inequality_beta} and \eqref{eq:inequality_L}.

Considering a starting point between the most inner and outermost tube, i.e., $\beta_k = -L_k \beta_{k, \mathcal{U}}$ for the $k\textsuperscript{th}$ tube with $1 < k < N$, this would lead to two patterns -- the ascending pattern \eqref{eq:beta_i_induction_most_outer} and the faulty descending pattern.
Hence, any of the $N-2$ remaining matrices cannot guarantee that \eqref{eq:inequality_beta} holds.
In conclusion, $\boldsymbol{M}_\mathcal{B}$ is the only linear mapping that utilizes the stated idea of successive selecting $\beta_k$ and considers both inequalities.

\section{Application to Joint Space Sampling}

In this section, we compare and contrast four rejection sampling methods with a sampling method utilizing the mapping $M_\mathcal{B}$.
For evaluation, we use a Monte Carlo method to determine the distribution and aggregate each rejection sampling method's average time and success rate.
A unit distribution is used of each $\beta_i$, where the three tubes lengths are set to \SI{100}{mm}, \SI{150}{mm}, and \SI{200}{mm}.

\subsection{Rejection Sampling via Branching}
\label{sec:rejection_sampling_via_branching}

Rejection sampling is the most common approach to sample $\beta_i$ subject to \eqref{eq:inequality_beta} and \eqref{eq:inequality_L}.
All values of the $k\textsuperscript{th}$ sample are sampled independently from a uniform distribution, i.e.,
\begin{align}
    \boldsymbol{\beta}^{\left(k\right)} \leftarrow \diag{\left(-L_i\right)}\boldsymbol{\beta}_{\mathcal{U}}^{\left(k\right)} \,
    \label{eq:sampling_vanilla}
\end{align}
where $\diag{\left(-L_i\right)}$ is short for $\diag{\left(-L_1, -L_2, -L_3\right)}$.
Afterward, both inequalities are checked for the $k\textsuperscript{th}$ sample, and, if necessary, a resampling approach is applied.
This step is repeated until \eqref{eq:inequality_beta} and \eqref{eq:inequality_L} hold.

To achieve a desired sample size, we consider four methods with different resampling approaches.
If the $k\textsuperscript{th}$ sample $\boldsymbol{\beta}^{\left(k\right)}$ does not satisfy both inequalities \eqref{eq:inequality_beta} and \eqref{eq:inequality_L}, we
\begin{itemize}
    \item[(a)] re-sample all values,
    \item[(b)] re-sample while keeping $\beta_1 = -L_1\beta_{1, \mathcal{U}}$,
    \item[(c)] re-sample while keeping $\beta_2 = -L_2\beta_{2, \mathcal{U}}$, or
    \item[(d)] re-sample while keeping $\beta_3 = -L_3\beta_{3, \mathcal{U}}$.
\end{itemize}
The first approach, i.e., (a), is the most common variant in the literature.
The other methods can be considered as a special case, where a specific joint value is preset.

\subsection{Direct Sampling with Affine Transformation Matrix}
\label{sec:sampling_via_M_B}

Another approach is to map a sampled point $\boldsymbol{\beta}_{\mathcal{U}}$ from a uniform cube $\mathcal{U}^3$ using $\boldsymbol{M}_\mathcal{B}$.
As proven in Sec.~\ref{sec:uniqueness_proof}, all samples fulfill \eqref{eq:inequality_beta} and \eqref{eq:inequality_L}.
For the $k\textsuperscript{th}$ sample, we can utilize
\begin{align}
    \boldsymbol{\beta}^{\left(k\right)} = \boldsymbol{M}_\mathcal{B}\boldsymbol{\beta}_{\mathcal{U}}^{\left(k\right)}
    \label{eq:sampling_M_B}
\end{align}
to sample a valid joint values $\beta_i^{\left(k\right)}$ using \eqref{eq:M_B}.

\subsection{Evaluation and Results}

\begin{figure}
    \centering
    \includegraphics[width=0.95\columnwidth]{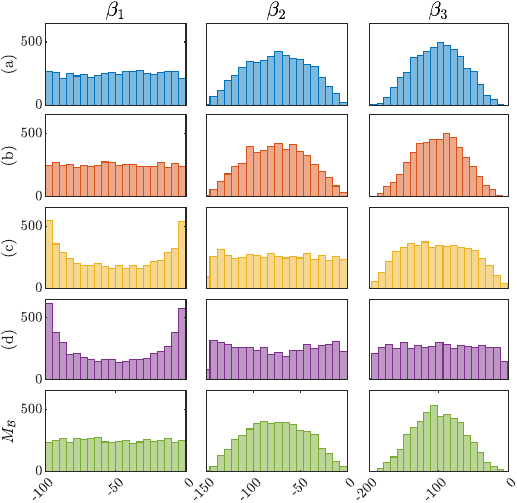}
    \vspace*{-0.75em}
    \caption{
        Distributions for different sampling method.
        The columns refer to the respective sampled joint values $\beta_i$, whereas the rows refer to the used sampling method. The rejection sampling methods (a) to (d) are described in Sec.~\ref{sec:rejection_sampling_via_branching}. The direct sampling using $\boldsymbol{M}_\mathcal{B}$ is described in Sec.~\ref{sec:sampling_via_M_B}.
    }
    \label{fig:distribution}
    \vspace*{-1.5em}
\end{figure}

Figure~\ref{fig:distribution} shows the distribution of each sampling method for \num{5000} samples.
The sample population is collected by running each method five times for \num{1000} samples.
Table~\ref{tab:sampling_comparision} lists the wall-clock time and success rate.
While a Matlab script on a consumer PC is used to generate the data, the relative increase of the wall-clock time is relevant for an implementation-independent comparison.

\begin{table}
    \caption{
        Wall-clock time and success rate for different sampling methods.
        The stated factor is the normalized wall-clock time w.r.t. the direct sampling method.
        Based on a maximal resampling attempt of \num{1000}, the fail rate indicates how often a sampling method fails to converge to a valid sample w.r.t all successful sampling attempts.
        Note that the complement of the success rate is the rate with which the sampled values are not valid and must be resampled.
    }
    \label{tab:sampling_comparision}
    \vspace*{-.75em}
    \centering
    \begin{tabular}{r r r r r} 
        \toprule
        method & time in \SI{}{ms} & factor & fail rate & success rate \\
		\cmidrule(r){1-1}
		\cmidrule(lr){2-2}
		\cmidrule(lr){3-3}
		\cmidrule(lr){4-4}
		\cmidrule(l){5-5}
        (a) & \num{58.8 \pm 3.6}  & \num{14.1} & N/A & \SI{8.4}{\%} \\
        (b) & \num{80.4 \pm 3.7}  & \num{19.3} & \SI{0}{\%} & \SI{8.3}{\%} \\
        (c) & \num{189.1 \pm 10.5}  & \num{45.3} & \SI{0.5}{\%} & \SI{3.2}{\%} \\
        (d) & \num{614.5 \pm 81.3}  & \num{147.3} & \SI{5.5}{\%} & \SI{1}{\%} \\
        $\boldsymbol{M}_\mathcal{B}$ & \num{4.2 \pm 0.3}  & 1 & N/A & \SI{100}{\%} \\
		\cmidrule(r){1-1}
		\cmidrule(lr){2-2}
		\cmidrule(lr){3-3}
		\cmidrule(lr){4-4}
		\cmidrule(l){5-5}
        $\boldsymbol{M}_\mathcal{B}$ (vectorized)& \num{0.3 \pm 0.3}  & \num{0.08} & N/A & \SI{100}{\%} \\
        \bottomrule
    \end{tabular}
\end{table}

The four sampling methods (a)-(d), branching  -- implemented via if-else statement -- and resampling -- implemented as a while loop -- are used, causing larger wall-clock time than the direct sampling method, which is at least two magnitudes faster.
On the contrary, our sampling method using $\boldsymbol{M}_\mathcal{B}$ is a branchless approach and offers three main advantages.
First, the success rate is \SI{100}{\%}.
Second, the execution is fast as it only relies on matrix multiplication with complexity of $O(N(N+1)/2)$.
Third, the wall-clock time can be further improved by vectorizing \eqref{eq:sampling_M_B}, where all samples can be computed in one step, which is impossible with other sampling methods.

For the simple and most commonly used approach, the success rate of the resampling approach (a) can be computed directly.
Using \eqref{eq:M_B_det}, success rate is equivalent to 
\begin{align}
    %
    \small 
    \dfrac{|\det \boldsymbol{M}_\mathcal{B}|}{\prod_{i=1}^N L_{i}} = \prod_{i=2}^N \dfrac{L_{i} - L_{i-1}}{L_{i}} = \prod_{i=2}^N \left(1 - \dfrac{L_{i-1}}{L_{i}}\right) < 1,
    \label{eq:success_rate}
\end{align}
being the ratio of the volume.
The theoretical value of \SI{8.3}{\%} is consistent with the empirical result \SI{8.4}{\%} listed in Table~\ref{tab:sampling_comparision}.
From \eqref{eq:success_rate}, each factor depends on the ratio of adjacent tubes and is smaller than one.
These facts reveal two characteristics.
First, the similar the lengths of the adjacent tubes are, the lower the success rate.
Second, the more nested tubes, the lower the success rate.
As a consequence, the lower \eqref{eq:success_rate}, the larger the wall-clock time.

Note the relation between \eqref{eq:beta_i_induction_most_outer_first_step} and the sampling method \eqref{eq:sampling_vanilla} with (b).
From that, it is no surprise that the unbiased sampling method (a), sampling method (b), and the sampling method using $\boldsymbol{M}_\mathcal{B}$ converges to similar distributions.
Looking into $\boldsymbol{M}_\mathcal{B}$ and the pattern \eqref{eq:beta_i_induction_most_outer} leading to \eqref{eq:M_B}, we can see that $\beta_i$ is the weighted sum of $i$ number of different independent scaled uniform distributions.
Therefore, the distribution of each $\beta_i$ relates to an Irwin-Hall distribution -- also known as the uniform sum distribution -- explaining the characteristics of distributed joint values for the three sampling methods.
This effect is clearly noticeable with a higher sample size, i.e., in \cite{GrassmannBurgner-Kahrs_et_al_IROS_2022}, the distributions shows a clear uniform, triangular, or bell-shaped distribution.
By contrast, the sampling methods (c) and (d) can result in a U-shaped distribution.
Note that the higher the value for the maximum resampling attempt, see Table~\ref{tab:sampling_comparision}, the more pronounced the U-shaped distribution for $\beta_2$ using (d).
These U-shaped characteristics can be seen in \cite{Burgner-KahrsWebster_et_al_IROS_2014} too.
Moreover, the wall-clock time of (b) and (c) is significantly higher, while the success rate is low.
Further note that we added an upper bound on the resampling attempts, see Table~\ref{tab:sampling_comparision}, such that (a), (b), and (c) terminate after a finite number of steps, otherwise the wall-clock time would have increased further.
This directly translates to downstream methods that use a resampling approach with preset joint values.

\section{Application to Workspace Estimation}

We will discuss an observation before applying our sampling method to workspace estimation.
The observation relates to the sampling of two variables to cover an area.
After presenting a solution to the observation, we briefly state a method to exploit rotational symmetries.
Finally, this section ends with a comparison of workspace estimation given different sampling methods.

\subsection{Translation and the Square Root Transformation}

\begin{figure}[tb]
    \centering
    \includegraphics[width=\columnwidth]{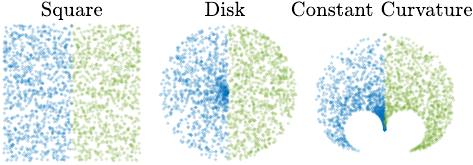}
    \vspace*{-1.5em}
    \caption{
    Affecting of distribution on two-dimensional workspace of toy examples.
    Two variables are randomly drawn from uniform distributions and used afterward to generate the workspace for a square, a disk, and a single-segment constant curvature continuum robot.
    Each workspace is divided into two areas -- (blue) vanilla approach and (green) proposed approach.
    The blue sampled points are linearly transformed, whereas, for the green sampled points, with the exception of the squared workspace, the additional transformation \eqref{eq:beta_sampling_area} is used for the translational variable.
    As can be seen from the realized distribution, the green sampled points yield a more homogeneous distribution.
    }
    \vspace*{-1em}
    \label{fig:sampling_area}
\end{figure}

We start with toy examples as motivation.
To cover an area, two random variables are required.
For the simple case of a rectangle, sampling both independent dimensions with uniform distribution results in expected distributed points in the area.
In contrast, when sampling a disk, the points are undesirably distributed, see Fig.~\ref{fig:sampling_area}.
Moreover, a planar workspace of a continuum robot with constant curvature kinematics and variable length is closely related to a circular area.
With no surprise, it shows a similar effect, see Fig.~\ref{fig:sampling_area}.

For the disk, one approach is to derive the Jacobian determinant of the transformation between the Cartesian and polar coordinate \cite{Dunbar1_CMJ_1997, EisenbergSullivan_AMM_2000}.
This leads to a simple solution for the disk, where the radius is determined by sampling its square.
Adapting this solution for $\beta_{i, \mathcal{U}}$ related to the tube length, this leads to
\begin{align}
    \left(\beta_{i, \mathcal{U}}\right)^2 = \mathcal{U}\left[0, 1\right].
    \label{eq:beta_sampling_area}
\end{align}
Afterward, the mapping \eqref{eq:M_B} can be used.
Note that, for the disk, the angle is independent of the length variable. However, this is not the case for the constant curvature.
We leave the proper sampling of the curvature for future work.

\subsection{Workspace Estimation}

To evaluate our proposed approach, we determine the workspace of two CTCRs.
We compare the sampling method stated in \eqref{eq:sampling_vanilla} with the resampling approach (a), and \eqref{eq:sampling_M_B}.
Both sampling methods are used with and without \eqref{eq:beta_sampling_area}.
For the former, $\boldsymbol{\beta}_{\mathcal{U}}^{\left(k\right)} = \sqrt{\mathcal{U}^N\left[0, 1\right]}$ is used in place of $\boldsymbol{\beta}_{\mathcal{U}}^{\left(k\right)} = \mathcal{U}^N\left[0, 1\right]$.
To create each dataset, \num{125000} tip positions are sampled for each sampling method and CTCR.
To compute the tip position, we use a static model \cite{RuckerJonesWebster_TRO_2010} with Poisson's ratio of \num{0.3}, Young's Modulus of \SI{50}{GPa}, and the geometrical parameters are stated in \cite{Burgner-KahrsWebster_et_al_IROS_2014} and \cite{GrassmannBurgner-Kahrs_et_al_IROS_2022}.

\begin{figure}
    \centering
    \includegraphics[width=0.95\columnwidth]{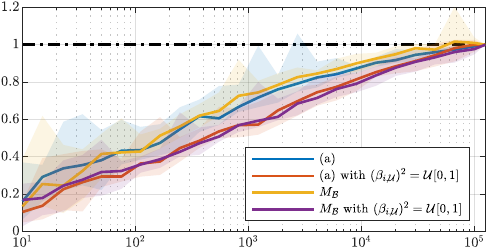}
    \\[1ex]
    \includegraphics[width=0.95\columnwidth]{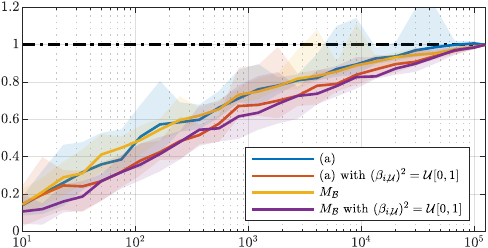}
    \vspace*{-0.5em}
    \caption{
    Convergence of workspace estimation in percent over the number of samples.
    The top and bottom plot show the workspace estimation for a CTCR with geometrical parameters stated in \cite{GrassmannBurgner-Kahrs_et_al_IROS_2022} and \cite{Burgner-KahrsWebster_et_al_IROS_2014}, respectively.
    The solid lines are the median values being limited by the minimum and maximum estimates.
    Ten different permutations of the ordered dataset are used to compute the median, minimum, and maximum area.
    Note that a new sampled point between two existing points at the boundary can reduce the computed area by the triangular area spanned by those three points and, therefore, the dashed \SI{100}{\%} mark can be exceeded.
    Further note that the workspace estimation using $\boldsymbol{M}_\mathcal{B}$ has a faster overall time of convergence.
    }
    \label{fig:convergence_workspace}
    \vspace*{-1.5em}
\end{figure}

The sampled tip positions are mapped onto a plane via as the workspace of the CTCR exhibits rotational symmetry.
This can be achieved either by plotting the distance between a point and the symmetry axis against the symmetry axis, or by rotation each point into a plane.\footnote{
By rotating all tip positions onto one plane, we like to call this the “Lantern method” as it reminds us of the opening and unfolding of a paper lantern.
This method \cite{BergelesDupont_et_al_T-RO_2015, NohLee_IROS_2016} is a simple yet useful simplification.}
Afterwards, a boundary (\textit{boundary} in MATLAB with shrink factor set to one) of the set of points is found and the area (\textit{polyarea} in MATLAB) is computed.
Figure~\ref{fig:convergence_workspace} shows the results of the workspace estimation.
Table~\ref{tab:workspace_comparision} lists the estimated area as well as the closeness.
The closeness is measured by the Euclidean distance between all projected tip positions.

While the absolute value reported in Table~\ref{tab:workspace_comparision} is less descriptive, the relative differences between sampling methods with and without \eqref{eq:beta_sampling_area} show that the mean and median distance are smaller and less spread according to the standard deviation.
From this, it can be concluded that using \eqref{eq:beta_sampling_area} can distributed the sampled tip position more evenly.
However, at the expense of a smaller estimated workspace.
From Fig.~\ref{fig:convergence_workspace}, the sampling method without \eqref{eq:beta_sampling_area} converges faster to the final area at the expense of less homogeneous distributed samples in the area.
Moreover, as expected, \eqref{eq:sampling_vanilla} with (a) re-sampling approach and \eqref{eq:sampling_M_B} perform similarly.
The exception here is the workspace estimation, where \eqref{eq:sampling_M_B} with \eqref{eq:beta_sampling_area} tends to generate a smaller area.

\begin{table}
    \caption{
        Area and closeness of sampled workspaces.
    }
    \label{tab:workspace_comparision}
    \vspace*{-0.5em}
    \centering
    \begin{tabular}{r r rr rr} 
        \toprule
        && \multicolumn{2}{N}{closeness} \\
		\cmidrule(lr){3-4}
        method & $\sqrt{\mathcal{U}^N}$ & mean & median$^1$ & area & CTCR$^2$ \\
		\cmidrule(r){1-1}
		\cmidrule(lr){2-2}
		\cmidrule(lr){3-3}
		\cmidrule(lr){4-4}
		\cmidrule(lr){5-5}
		\cmidrule(l){6-6}
        (a) & \checkNo & \num{0.056 \pm 0.237} & \num{0.052} & \num{0.0175} & \cite{GrassmannBurgner-Kahrs_et_al_IROS_2022}\\
        (a) & \checkYes & \num{0.038 \pm 0.195} & \num{0.032} & \num{0.0144} & \cite{GrassmannBurgner-Kahrs_et_al_IROS_2022} \\
        $\boldsymbol{M}_\mathcal{B}$ & \checkNo  & \num{0.056 \pm 0.237} & \num{0.053} & \num{0.0171} & \cite{GrassmannBurgner-Kahrs_et_al_IROS_2022}\\
        $\boldsymbol{M}_\mathcal{B}$ & \checkYes  & \num{0.040 \pm 0.200} & \num{0.034} & \num{0.0134} & \cite{GrassmannBurgner-Kahrs_et_al_IROS_2022}\\
		\cmidrule(r){1-1}
		\cmidrule(lr){2-2}
		\cmidrule(lr){3-3}
		\cmidrule(lr){4-4}
		\cmidrule(lr){5-5}
		\cmidrule(l){6-6}
        (a) & \checkNo & \num{0.049 \pm 0.221} & \num{0.045} & \num{0.0131} & \cite{Burgner-KahrsWebster_et_al_IROS_2014}\\
        (a) & \checkYes & \num{0.037 \pm 0.193} & \num{0.032} & \num{0.0115} & \cite{Burgner-KahrsWebster_et_al_IROS_2014} \\
        $\boldsymbol{M}_\mathcal{B}$ & \checkNo  & \num{0.049 \pm 0.221} & \num{0.045} & \num{0.0131} & \cite{Burgner-KahrsWebster_et_al_IROS_2014}\\
        $\boldsymbol{M}_\mathcal{B}$ & \checkYes  & \num{0.037 \pm 0.192} & \num{0.032} & \num{0.0103} & \cite{Burgner-KahrsWebster_et_al_IROS_2014}\\
        \bottomrule
    \end{tabular}
	\hspace*{5mm}
	\begin{itemize}
		\item[$^1$]\hspace{-2.5mm} Median of the medians is computed instead of the median directly since $S = \num{125 000}$ samples require to compare $S(S-1)/2$ distances, which is too big -- \SI{58.2}{GB} -- to fit in the memory of our machine.
		\item[$^2$]\hspace{-2.5mm} Reference to the parameters of the CTCR. For \cite{Burgner-KahrsWebster_et_al_IROS_2014}, the \textit{Robot 1} defined in Table~I is used.
	\end{itemize}
\end{table}

\section{Application to Control}

\begin{figure}
    \centering
    \includegraphics[width=\columnwidth]{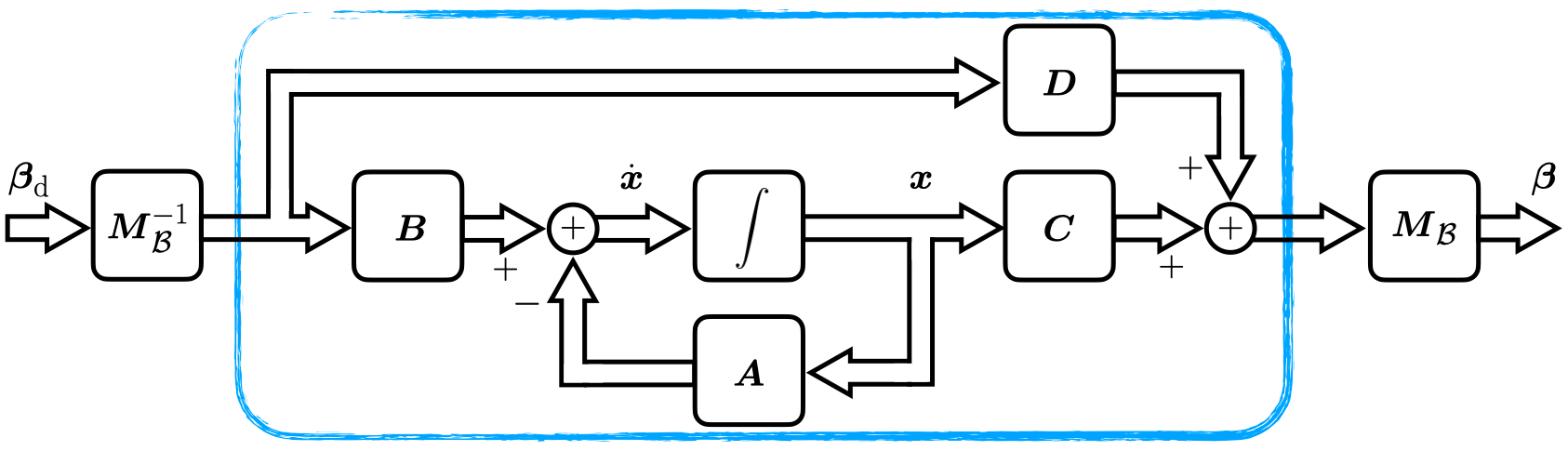}
    \caption{
    Block diagram of the closed-loop system in state-space.
    The system encased in blue represents is a simple PT\textsubscript{1} model of the simplified motor dynamics controlled by a PI controller.
    This system must account for the two interdependent inequalities \eqref{eq:inequality_beta} and \eqref{eq:inequality_L}.
    By using $M_\mathcal{B}$, they become independent box constraints that are easier to handle.
    }
    \label{fig:blockdiagram}
\end{figure}

In this section, we will incorporate $M_\mathcal{B}$ to a control application.
This is motivated by the fact that a closed-loop system with a simple control for $\beta_i$ can output values that do not satisfy \eqref{eq:inequality_beta} and \eqref{eq:inequality_L}.
This is possible even if all inputs are valid desired $\beta_{i, \mathrm{d}}$.
For the sake of simplicity, a first-order proportional delay element (PT\textsubscript{1}) model and a PI controller with $K_{P, i}$ and $K_{I, i}$ gains for the $i\textsuperscript{th}$ tube are used for the closed-loop control.
This leads to a simple linear and time-invariant system.
The block diagram is shown in Fig.~\ref{fig:blockdiagram}.

\subsection{Vanilla Approach}

The vanilla approach treats each $\beta_i$ independently.
Therefore, the block diagram looks similar to the one shown in Fig.~\ref{fig:blockdiagram}, where the additional transformation using $\boldsymbol{M}_\mathcal{B}$ is set to the identity matrix.
The matrices of the state equation are
\begin{align}
	\boldsymbol{A}
	=
	\begin{bmatrix}
		\diag{\left(-K_{P, i} - 1\right)} & \diag{\left(K_{I, i}\right)} \\
		-\boldsymbol{I}_{3 \times 3} & \boldsymbol{0}_{3 \times 3}\\
	\end{bmatrix}
	\label{eq:A}
\end{align}
and $\boldsymbol{B} = \left[\diag{\left(K_{P, i}\right)}, \boldsymbol{I}_{3 \times 3}\right]\transpose$.
The output equations are given by $\boldsymbol{C} = \left[\boldsymbol{I}_{3 \times 3}, \boldsymbol{0}_{3 \times 3}\right]$ and $\boldsymbol{D} = \boldsymbol{0}_{3 \times 3}$.

\subsection{Transforming the State Space Representation}

A rather novel approach utilizing \eqref{eq:M_B} is to transform the state space vector as shown in Figure~\ref{fig:blockdiagram}.
Using block diagram algebra, the state equation changes to
\begin{align}
    \boldsymbol{\widehat{A}}
    &=
	\left[\diag{\left(\boldsymbol{M}_\mathcal{B}, \boldsymbol{M}_\mathcal{B}\right)}\right] \boldsymbol{A} \left[\diag{\left(\boldsymbol{M}_\mathcal{B}^{-1}, \boldsymbol{M}_\mathcal{B}^{-1}\right)}\right],
	\label{eq:A_prime}
\end{align}
where $\diag{\left(\boldsymbol{M}_\mathcal{B}, \boldsymbol{M}_\mathcal{B}\right)}$ and $\diag{\left(\boldsymbol{M}_\mathcal{B}^{-1}, \boldsymbol{M}_\mathcal{B}^{-1}\right)}$ are block diagonal matrices whose diagonal contains blocks of $\boldsymbol{M}_\mathcal{B}$ and $\boldsymbol{M}_\mathcal{B}^{-1}$, respectively.
From \eqref{eq:A_prime}, we get $\boldsymbol{M}_\mathcal{B}\diag{\left(-K_{P, i} - 1\right)}\boldsymbol{M}_\mathcal{B}^{-1}$ resulting in
\begin{align}
	\begin{bmatrix}
		- K_{P, 1} - 1 & 0 & 0 \\
		K_{P, 2} - K_{P, 1} & - K_{P, 2} - 1 & 0 \\
		K_{P, 2} - K_{P, 1} & K_{P, 3} - K_{P, 2} & - K_{P, 3} - 1\\
	\end{bmatrix},
	\label{eq:diag_Kp_prime}
\end{align}
whereas the upper right block, i.e., $\boldsymbol{M}_\mathcal{B}\diag{\left(K_{I, i}\right)}\boldsymbol{M}_\mathcal{B}^{-1}$, is 
\begin{align}
	\begin{bmatrix}
		K_{I, 1} & 0 & 0 \\
		K_{I, 1} - K_{I, 2} & K_{I, 2} & 0 \\
		K_{I, 1} - K_{I, 2} & K_{I, 2} - K_{I, 3} & K_{I, 3}\\
	\end{bmatrix}.
	\label{eq:diag_Ki_prime}
\end{align}
The matrices $\boldsymbol{\widehat{C}} = \boldsymbol{C}$ and $\boldsymbol{\widehat{D}} = \boldsymbol{D}$ stay unchanged, while 
\begin{align}
    \boldsymbol{\widehat{B}}
    = 
	\begin{bmatrix}
		\boldsymbol{M}_\mathcal{B}\diag{\left(K_{P, i}\right)}\boldsymbol{M}_\mathcal{B}^{-1}\\
		\boldsymbol{I}_{3 \times 3}\\
	\end{bmatrix}
    \label{eq:B_prime}
\end{align}
is the transformed $\boldsymbol{B}$, where $\boldsymbol{M}_\mathcal{B}\diag{\left(K_{P, i}\right)}\boldsymbol{M}_\mathcal{B}^{-1}$ has a similar appearance to \eqref{eq:diag_Kp_prime}, only it negative and without ones.

\subsection{Advantages}

While inequalities \eqref{eq:inequality_beta} and \eqref{eq:inequality_L} can be incorporated and considered in an optimal control scheme, using $\boldsymbol{M}_\mathcal{B}$, the inequalities simplify to box constraints.
Hence, the controller to be synthesized can be more computationally efficient compared to an optimal control scheme.
In other words, transforming the state space representation leads to a system, which can be treated as a linear time-invariant system.
Therefore, a larger variety of methods are at the disposal of the user.
Furthermore, nonlinear elements such as saturation and anti-windup elements can be integrated with ease.

Another benefit is the analysis of such transformed system, which can be straightforward.
For instance, for \eqref{eq:A_prime} of the transformed system, we can see that the constraints
\begin{align}
	K_{P, i} \leq K_{P, i+1} \quad\text{and}\quad K_{I, i+1} \leq K_{I, i}
\end{align}
for \eqref{eq:diag_Kp_prime} and \eqref{eq:diag_Ki_prime}, respectively, are necessary otherwise the system becomes a non-minimum phase system.
Therefore, an undesirable set of controller gains can make the closed-loop system unstable.
This insight gives rise to possible challenges in control synthesis for CTCRs, when using a branching method similar to Sec.~\ref{sec:rejection_sampling_via_branching}.

\section{Discussion}

Using $\boldsymbol{M}_\mathcal{B}$ has been shown benefical for machine leaning application \cite{GrassmannBurgner-Kahrs_RSS_2019, GrassmannBurgner-Kahrs_et_al_IROS_2022, IyengarStoyanov_et_al_RA-L_2023}.
In this paper, we touch on the application in control, workspace estimation, and, in a more general sense, in sampling.
While the results are preliminary, they show promising future direction utilizing the advantage of $\boldsymbol{M}_\mathcal{B}$ being the transformation of \eqref{eq:inequality_beta} and \eqref{eq:inequality_L} to $N$ box constraints.
This opens the possibility to use branchless approaches offering several advantages.
Among others, it reduces the time spent on branch prediction, resulting in faster execution. 
This is particularly beneficial for real-time applications that require quick responses, such as control systems. 
Additionally, in computational hungry applications like tube design or online learning, where reducing time is crucial to ensure efficient resource utilization.
Another benefit of utilizing $\boldsymbol{M}_\mathcal{B}$ is that it gives the possibility to shape distributions to any desired distributions using the inverse transform sampling.

We acknowledge that other representations and notations exist.
For example, in \cite{LyonsWebsterAlterovitz_IROS_2009, LyonsWebsterAlterovitz_ICRA_2010}, the translation is defined by the segment length of overlapping tubes, i.e., $\l_i = \beta_i + L_i - \left(\beta_{i-1} + L_{i-1}\right) \geq 0$.
This definition considers only the manipulator side, i.e., \eqref{eq:inequality_L}, which has its merits for algorithm in simulation.
However, the algorithm applied to real hardware must account for \eqref{eq:inequality_beta}, which reflects the actuation side.
Moreover, $\l_i$ either needs to be measured or related to $\beta_i$, which does not solve the inequalities in the first place.
Furthermore, indexing starts either with the outermost tube, e.g., \cite{WebsterRomanoCowan_T-RO_2009, BaranTamadazte_IROS_2017, GrassmannBurgner-Kahrs_et_al_IROS_2022}, or with the most inner tube, e.g., \cite{YangMeng_et_al_Access_2019, GrassmannBurgner-Kahrs_RSS_2019}.
We advocate using the former, showing advantages in formalizing patterns, sequence, and proofs as used in Sec.~\ref{sec:uniqueness_proof}.
In any case, caution is advised when comparing formulae and algorithms using different notations.
For example, the upper triangular matrix in \cite{GrassmannBurgner-Kahrs_RSS_2019} is the low triangular matrix \eqref{eq:M_B} and vise versa.

\section{Conclusions}

In this paper, we investigate the affine transformation matrix $\boldsymbol{M}_\mathcal{B}$ to mitigate the use of branching methods.
Its derivation and uniqueness are shown.
Use of this transformation has ripple effects on downstream approaches previously using sampling or branching.
We explored the application to sampling, workspace estimation, and control showing advantages in success rates, lowering the computational load, and interpretability of used methods.

\addcontentsline{toc}{section}{REFERENCES}
\bibliographystyle{IEEEtran}
\bibliography{IEEEabrv, references}

\end{document}